\documentclass[10pt,twocolumn,letterpaper]{article}
\usepackage{placeins}
\usepackage[camera]{cvpr}      %
\usepackage{float}

\definecolor{cvprblue}{rgb}{0.21,0.49,0.74}

\usepackage[pagebackref,breaklinks,colorlinks,allcolors=cvprblue]{hyperref}

\def\paperID{11453} %
\def\confName{CVPR}
\def\confYear{2025}

\title{DiffLoRA: Generating Personalized Low-Rank Adaptation Weights with Diffusion Models}

\author{Yujia Wu\textsuperscript{\rm 1} Yiming Shi\textsuperscript{\rm 1}, Jiwei Wei\textsuperscript{\rm 1} Chengwei Sun\textsuperscript{\rm 1} \\ Yang Yang\textsuperscript{\rm 1} Heng Tao Shen\textsuperscript{\rm 1}\\
\textsuperscript{\rm 1}University of Electronic Science and Technology of China\\
}

\begin{document}

\twocolumn[{
\maketitle
\begin{center}
    \captionsetup{type=figure}
    \includegraphics[width=\textwidth, keepaspectratio]{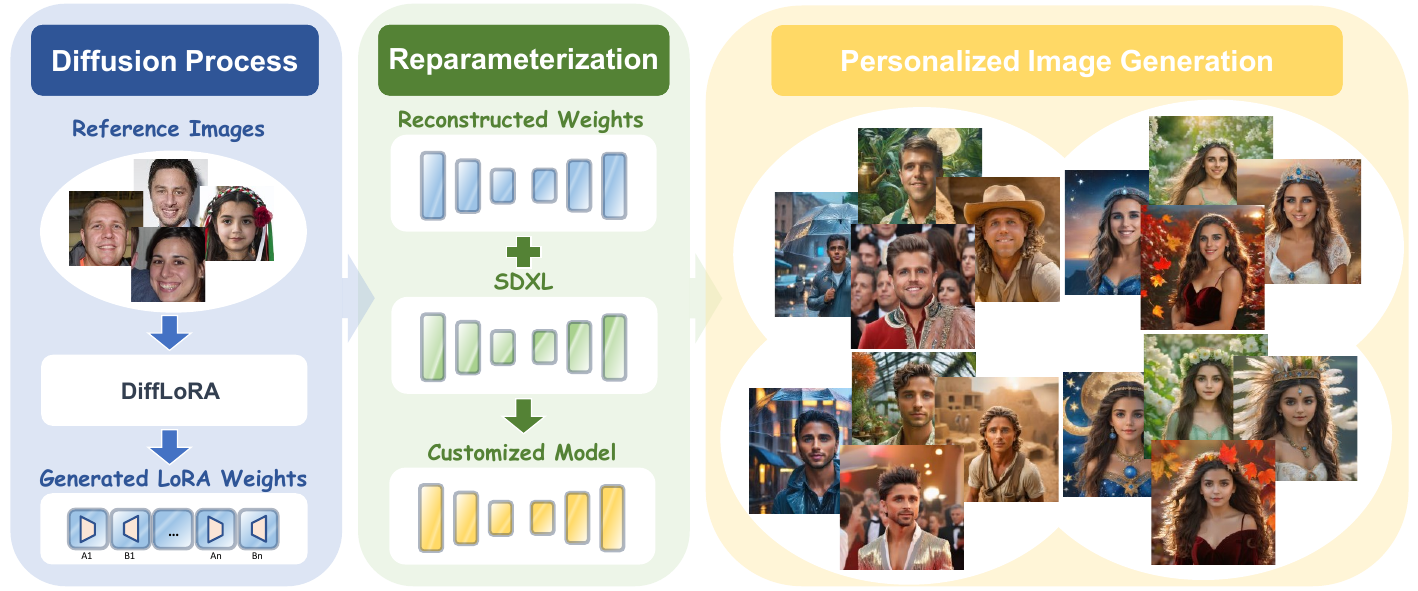}
    \captionof{figure}{DiffLoRA generates personalized LoRA weights based on reference images, enabling personalized image synthesis through directly merging these weights into the off-the-shelf text-to-image model.
    \label{fig:DiffLoRA}}
\end{center}
}]
\maketitle

\begin{abstract}

    Personalized text-to-image generation has gained significant attention for its capability to generate high-fidelity portraits of specific identities conditioned on user-defined prompts. Existing methods typically involve test-time fine-tuning or incorporating an additional pre-trained branch. However, these approaches struggle to simultaneously address efficiency, identity fidelity, and the preservation of the model's original generative capabilities. In this paper, we propose \textbf{DiffLoRA}, an efficient method that leverages the diffusion model as a hypernetwork to predict personalized Low-Rank Adaptation (LoRA) weights based on the reference images.~By incorporating these LoRA weights into the off-the-shelf text-to-image model, DiffLoRA enables zero-shot personalization during inference, eliminating the need for post-processing optimization. Moreover, we introduce a novel identity-oriented LoRA weights construction pipeline to facilitate the training process of DiffLoRA. The dataset generated through this pipeline enables DiffLoRA to produce consistently high-quality LoRA weights. Notably, the distinctive properties of the diffusion model enhance the generation of superior weights by employing probabilistic modeling to capture intricate structural patterns and thoroughly explore the weight space.~Comprehensive experimental results demonstrate that DiffLoRA outperforms existing personalization approaches across multiple benchmarks, achieving both time efficiency and maintaining identity fidelity throughout the personalization process.
\end{abstract}
    
\section{Introduction}
\label{sec:intro}

\hspace{1em} Recent significant advancements in large-scale text-to-image diffusion models have spurred extensive research into their customizability \cite{ddpm, rombach2022high, ddim, croitoru2023diffusion}.~A prominent example is human-centric customized image generation \cite{liuhyperhuman, ruiz2024hyperdreambooth, li2024photomaker, wang2024instantid}, which has garnered substantial attention due to its numerous applications, including personalized images with custom styles \cite{ren2022neural, cui2024idadapter, zhang2023sadtalker}, as well as controllable human image generation \cite{wang2018toward, zhou2022cross, humansd}. The core principle underlying these applications is to incorporate user-defined individuals into the generation process while maintaining identity consistency in the synthesized images.

To address this demand, various advanced methodologies have been developed for personalization in text-to-image synthesis. One significant approach is model fine-tuning with specific reference images, as outlined in works such as \cite{ruiz2023dreambooth, gal2022image, kumari2023multi, tewel2023key}. These methods can generate high-fidelity images while maintaining the original model's capabilities. However, they require substantial datasets and extensive training, consequently leading to processing times of 10 to 30 minutes per individual personalization, rendering them impractical for user-centric applications.~Conversely, studies like \cite{ye2023ip, wang2024instantid, li2024photomaker, xiao2023fastcomposer, he2024imagine} achieve personalization without test-time fine-tuning by incorporating an additional trainable condition branch. While these tuning-free methods provide a viable alternative, they often compromise the model's fidelity and versatility, struggling with maintaining identity fidelity and the original generative capabilities of the model \cite{zeng2024jedi}. Moreover, freezing the original diffusion model's weights to train an auxiliary learnable branch also incurs extra training cost and increases inference cost compared to the original diffusion model \cite{humansd}.

Notably, HyperDreambooth \cite{ruiz2024hyperdreambooth} provides an alternative solution to the aforementioned challenges by employing a transformer decoder as a hypernetwork to predict weights for personalization.~Nevertheless, constrained by the architectural limitations of the hypernetwork, HyperDreambooth can only predict a limited subset of Lightweight DreamBooth (LiDB) weights, resulting in inferior identity-fidelity and representation capabilities compared to the conventional LoRA weights. Consequently, after merging the LiDB weights into the text-to-image model, additional fast fine-tuning becomes inevitable for satisfactory personalization, imposing substantial computational overhead and lacking flexibility in practical applications.

Inspired by the remarkable generative capabilities of diffusion models and the effectiveness of LoRA \cite{hulora}, we propose DiffLoRA, an efficient latent diffusion-based~hypernetwork designed to generate~ready-to-use~personalized~LoRA weights. DiffLoRA consists~of an~autoencoder for LoRA weights and a diffusion transformer model specifically tailored for LoRA latent representations. Our method incorporates a Mixture-of-Experts-inspired (MoE-inspired) gate network \cite{shazeer2016outrageously} to synthesize comprehensive feature representations by combining facial and global characteristics from reference images. During inference, the~diffusion model leverages these mixed features as conditional inputs to guide the denoising process. By~reconstructing the generated LoRA latent representations with the decoder, we obtain the corresponding LoRA weights for specific identity adaptation in Stable Diffusion XL (SDXL) \cite{podell2023sdxl}. Such a latent diffusion-based hypernetwork enables predicting a larger parameter space compared to existing approaches \cite{ruiz2024hyperdreambooth, erkocc2023hyperdiffusion}. Additionally, the diffusion architecture facilitates cross-modality generation, achieving enhanced stability and superior results through iterative denoising \cite{ruan2023mm, metadiff, erkocc2023hyperdiffusion}. By directly incorporating the generated LoRA weights into SDXL, our approach achieves personalization without additional fine-tuning, while maintaining identity fidelity and preserving the original generative capabilities without introducing extra computational overhead.

Our DiffLoRA necessitates LoRA weights for multiple identities during training, thereby requiring the support of a specialized LoRA weights dataset. However, no standard dataset currently exists for LoRA weights of specific identities. To address this limitation, we develop an automated pipeline for constructing a LoRA weights dataset tailored for various identities. Our pipeline processes diverse images for each identity, encompassing various expressions, attributes, and scenes, and employs LoRA-DreamBooth \cite{von-platen-etal-2022-diffusers} to generate high-quality LoRA weights.

The summary of our contributions is as follows:
\begin{itemize}[leftmargin=1.2em, labelsep=0.6em]
    \item Through investigating cross-modal interactions between image representations and weight manifolds, we propose a novel paradigm to leverage diffusion models guided by reference images to generate LoRA weights for personalized image synthesis. This tuning-free approach enables high-fidelity personalized image generation without additional computational costs during inference.
    
    \item We introduce a new pipeline for constructing the LoRA weights dataset, which enhances DiffLoRA training and ensures the representation of a wide range of identities.
    
    \item Comprehensive experiments demonstrate that our method significantly outperforms existing state-of-the-art approaches in text-image consistency, identity fidelity, generation quality, and inference cost.
\end{itemize}

\section{Related Work and Motivation}
\label{sec:related_work}
\subsection{Low-Rank Adaptation (LoRA)} 
\hspace{1em} LoRA \cite{hulora} has emerged as an effective Parameter-Efficient Fine-Tuning (PEFT) approach by introducing trainable low-rank matrices alongside frozen pre-trained weights $W_0$. Specifically, LoRA decomposes the weight update $\Delta W$ into two low-rank matrices $A$ and $B$ for parameter adaptation. The~fine-tuned weight $W'$ can be mathematically represented as:
\begin{equation}
\begin{aligned}
W' = W_0 + \Delta W = W_0 + BA,
\end{aligned}
\label{eq_1}
\end{equation}
where $B \in \mathbb{R}^{m \times r}$, $A \in \mathbb{R}^{r \times n}$ with rank $r \ll \min(m, n)$. LoRA significantly reduces computational costs while maintaining comparable performance to full fine-tuning.~In \cref{3.1motivation}, we explore the inherent compatibility between LoRA weights and latent diffusion models (LDMs), which forms the theoretical foundation of our methodology.

\subsection{Versatile Applications of Diffusion Models} 
\hspace{1em} Diffusion probabilistic models \cite{ddpm,ddim} have emerged as a powerful generative framework, demonstrating superior quality and sample diversity compared to its precursors like GANs \cite{goodfellow2020generative} and VAEs \cite{kingma2013auto}. Building on this foundation, diffusion models have achieved remarkable success across multiple domains, with Stable Diffusion and its variants \cite{rombach2022high, podell2023sdxl, blackforestlabs2024, xiao2024omnigen} revolutionizing text-to-image generation, and DiffWave \cite{kong2020diffwave} pioneers the use of diffusion models for audio synthesis. Furthermore, diffusion models have shown great potential in parameter generation tasks. For instance, Learning to Learn \cite{peebles2022learning} enables single-step neural network optimization by training a diffusion model on 23 million checkpoint datasets. HyperDiffusion \cite{erkocc2023hyperdiffusion} leverages diffusion models to predict neural radiance field (NeRF) parameters. The study by \cite{wang2024neural, jin2024conditional} employ diffusion-based frameworks to generate adaptive parameters for diverse classification objectives. In contrast to these methods, our proposed solution, DiffLoRA, further extends the application of diffusion models to personalized LoRA weights generation.

\subsection{Personalization in Diffusion Models}
\hspace{1em} Personalized image generation using diffusion models has gained significant attention in recent research. Existing personalization approaches primarily fall into two categories.~Fine-tuning based methods,~such as DreamBooth \cite{ruiz2023dreambooth},~Textual Inversion \cite{gal2022image},~and~CustomDiffusion \cite{kumari2023multi}, achieve high-fidelity results through subject-specific model optimization, yet the computational overhead~limits their~practical~applications.~To~circumvent~time-consuming training, tuning-free~approaches~incorporate additional branches to inject identity information~during inference \cite{wang2024instantid, li2024photomaker, wei2023elite, valevski2023face0}.~However, these methods struggle to maintain a balanced trade-off between identity preservation and the model's inherent generative capabilities.~Recently, HyperDreambooth \cite{ruiz2024hyperdreambooth}~introduces~an~alternative solution through parameter generation, which addresses these personalization challenges but still necessitates an additional training phase, severely limiting its practical utility. Unlike HyperDreambooth, DiffLoRA directly generates and integrates LoRA weights from a reference image into SDXL without any requirement for retraining.

\section{Method}
\label{method}

\begin{figure*}[htbp]
    \centering
    \vspace{-1em}
    \includegraphics[width=\textwidth, keepaspectratio]{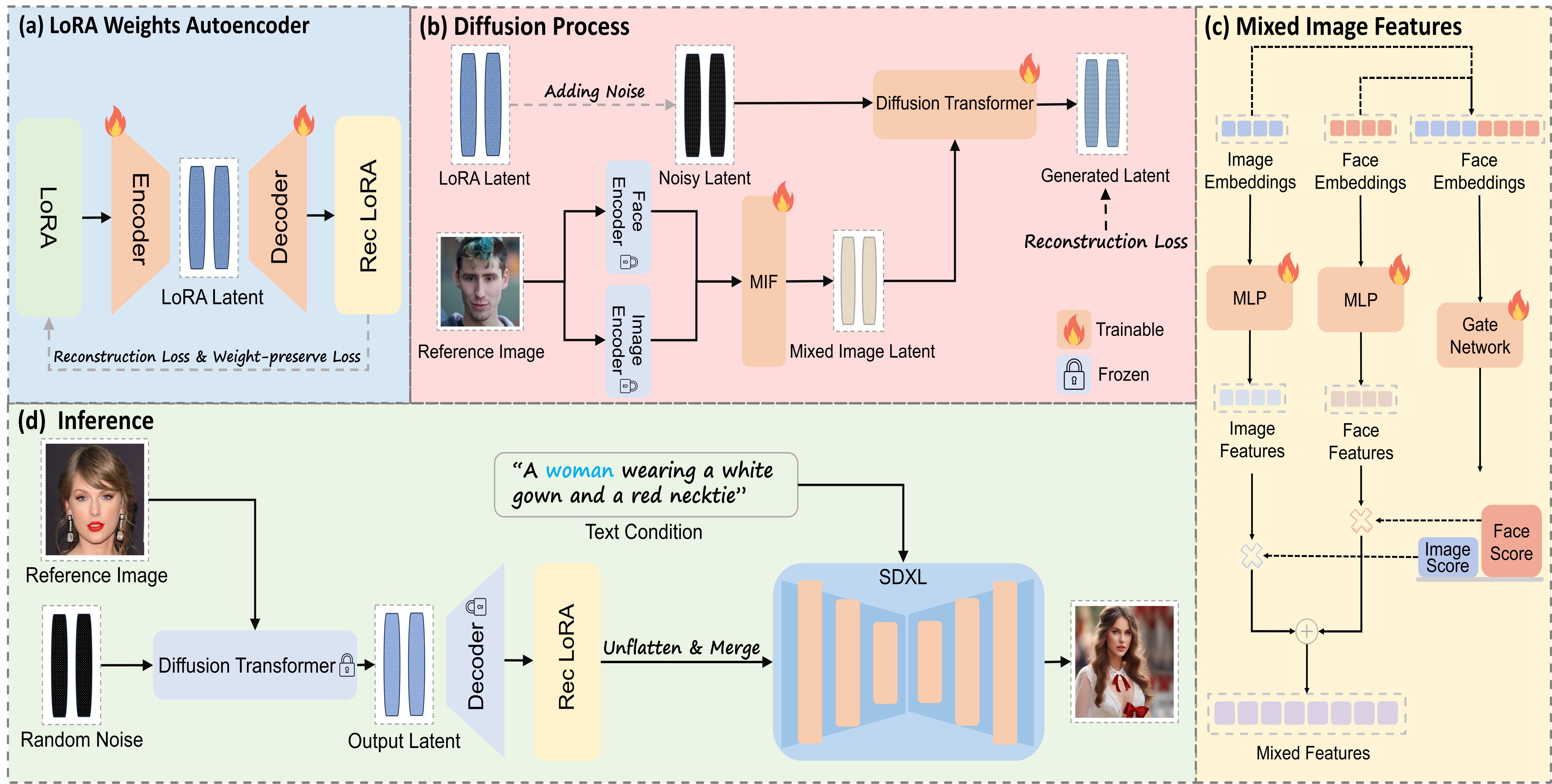}
    \vspace{-0.9em}
    \caption{Overview of DiffLoRA. We begin by encoding and reconstructing the LoRA weights using the LoRA Weights Autoencoder (LAE) to compress them into latent representations. In the diffusion process, these noisy latent representations are processed by a diffusion transformer conditioned on Mixed Image Features (MIF), integrating both facial and image features from reference images. During inference, the diffusion model takes random noise and reference images as input to generate personalized LoRA weights.}
    \vspace{-1em}
    \label{fig:train-inference}
\end{figure*}

\hspace{1em} Given reference images, DiffLoRA aims to efficiently generate LoRA weights by leveraging a LoRA weights autoencoder (LAE) combined with a diffusion model. This approach circumvents the extra computational costs associated with dual-branch architectures while achieving high-fidelity image generation through LoRA weights generation and reparameterization. We first explore the motivation behind generating LoRA weights with LDMs, highlighting their inherent compatibility (\cref{3.1motivation}). Following this, we detail the construction of the LAE for LoRA weights compression (\cref{LAE}) and describe our diffusion process that directly predicts encoded LoRA latent representations (\cref{diffusion process}). The Mixed Image Features (MIF) module is then introduced to integrate face and image features for guiding the denoising process (\cref{mif}). Lastly, we outline the dataset construction pipeline for generating~multi-identity LoRA weights (\cref{pipeline}). \cref{fig:train-inference} provides an overview of our training and inference process.

\subsection{Motivation}
\label{3.1motivation}
\hspace{1em} We explore the inherent compatibility between LoRA and LDMs through empirical analysis. Two key characteristics make LoRA particularly suitable for LDM-based parameter generation: the \textit{efficiency of low-rank structures} and the \textit{constrained distribution}.

The low-rank structures of LoRA significantly reduces the parameter space by decomposing the weight update $\Delta W$ into two low-rank matrices $A$ and $B$. This reduction from $m \times n$ to $(m + n) \times r$ parameters aligns well with LDM's efficient representation learning capability. Moreover, as observed in \cite{biderman2024lora}, LoRA weights exhibit a concentrated and stable distribution, which not only enhances generalization but also facilitates compression and reconstruction. These characteristics resonate with the core principle of LDMs, which operate in a compressed latent space.

\begin{figure}[t]
    \centering
    \includegraphics[width=\columnwidth]{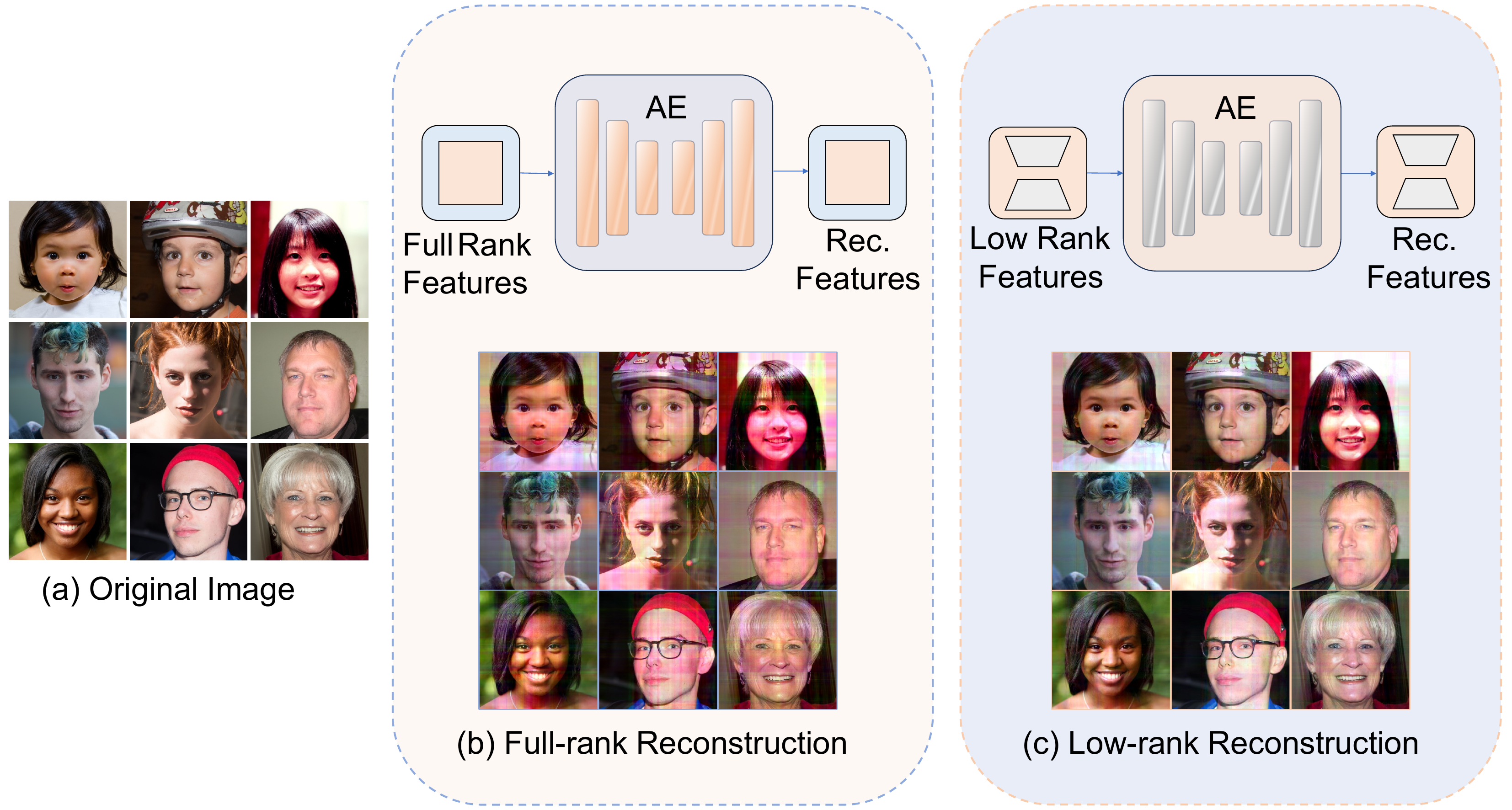}
    \caption{A toy experiment demonstrating superior compression and reconstruction performance of low-rank features.}
    \vspace{-1em}
    \label{fig:toy_experiment}
\end{figure}

To better demonstrate our theoretical claims, we design a toy experiment focusing on the compression and reconstruction of low-rank features. Following the methodology outlined by \cite{prasantha2007image}, we perform SVD \cite{klema1980singular} on face images from the FFHQ dataset \cite{FFHQ} and extract the low-rank matrices of the top few dimensions to obtain the low-rank features. This method leverages SVD's energy compaction property to obtain efficient low-rank representations \cite{guo2015efficient}. Subsequently, we train two autoencoders with identical architectures to reconstruct full-rank and low-rank features of the images, respectively. Both autoencoders aim to reconstruct their input features under the same training conditions. Once the autoencoder trained on low-rank features converges, we evaluate the reconstruction quality of both models using peak signal-to-noise ratio (PSNR) and structural similarity index (SSIM) metrics \cite{hore2010image}. The results, as shown in \cref{fig:toy_experiment}, demonstrate that the autoencoder reconstructing low-rank features achieves higher values in both PSNR and SSIM, indicating superior reconstruction quality. 

This empirical evidence substantiates that low-rank structures are inherently conducive to compression and reconstruction, demonstrating the fundamental compatibility between LoRA and LDMs. More details of the experiment are provided in the Supplementary Materials.

\subsection{LoRA Weights Autoencoder}
\label{LAE}

\hspace{1em} As shown in \cref{fig:train-inference}(a), the LoRA weights autoencoder (LAE) is designed to effectively compress and reconstruct the LoRA weights, specifically targeting the structural and informational correlations inherent in LoRA weights.

For structural features, LoRA weights present two distinct shapes: LoRA-A$ \in \mathbb{R}^{r \times n}$ and LoRA-B$ \in \mathbb{R}^{m \times r}$, necessitating a flattening operation. To maximize the preservation of structural information in the latent space, we directly flatten LoRA-B while transposing LoRA-A to $\mathbb{R}^{n \times r}$ before flattening. The resulting one-dimensional vectors are concatenated as input to the LAE. We then employ 1D convolutional layers as the primary compression layers to capture the structural features of the LoRA weights.

\begin{figure}[t]
    \centering
    \includegraphics[width=\columnwidth]{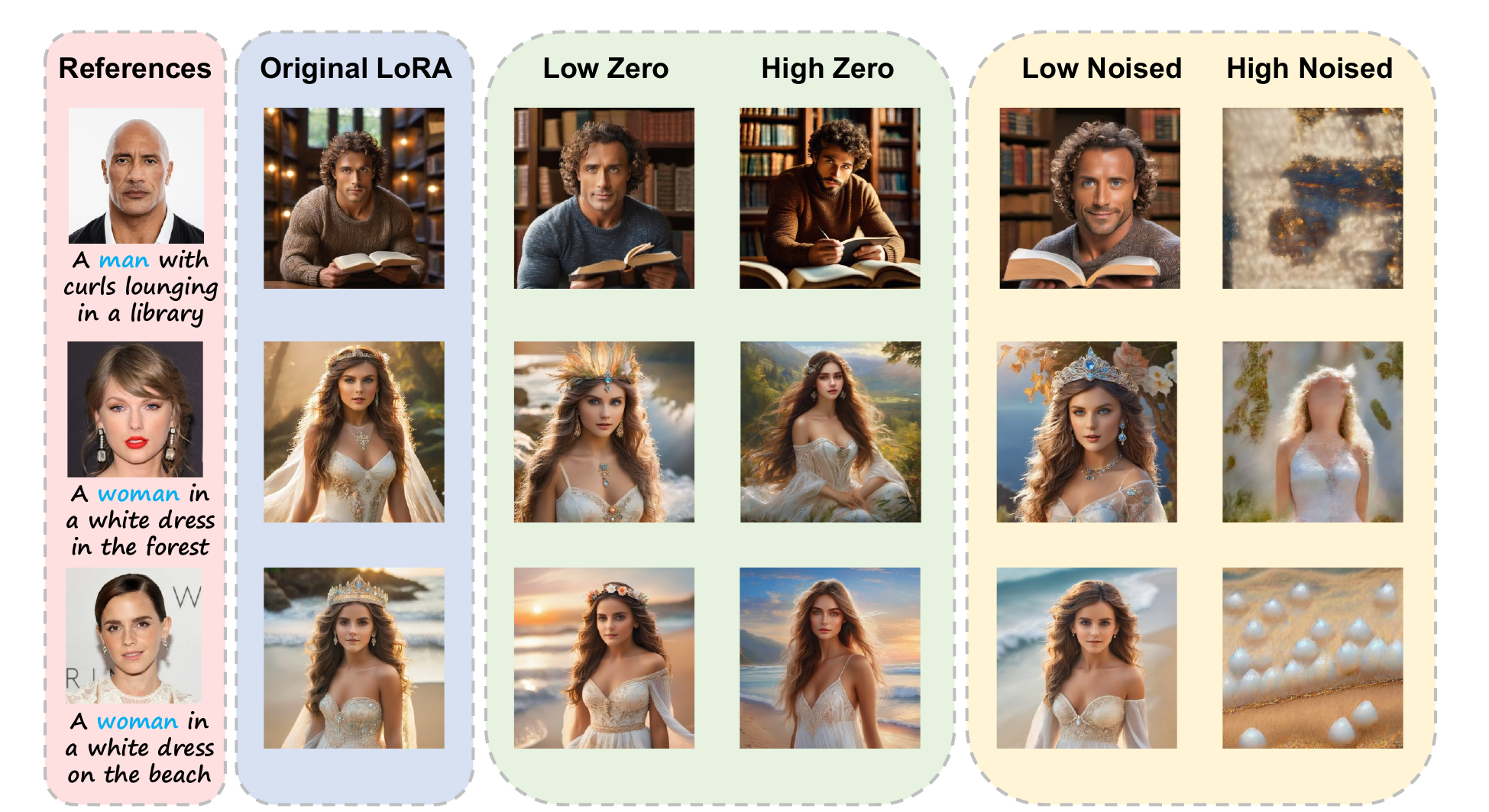}
    \caption{An example illustrating the importance of large LoRA weights in preserving identity information.}
    \label{fig:information_features}
\end{figure}

Concerning the informational features of LoRA weights, we conduct an in-depth analysis of the impact of different magnitudes of personalized LoRA weights on the final generated images. We sort the LoRA weights by their magnitude and divided them into four proportional groups: the first group representing the large weights and the last group representing the small weights. Subsequently, we either set the large weights and small weights to zero or perturb them with proportionally scaled noise. As illustrated in \cref{fig:information_features}, we found that manipulating small weights had minimal impact on the generated images while altering large weights significantly degraded identity fidelity.

This analysis demonstrates that large LoRA weights are more instrumental in encoding specific identity information. Leveraging this insight, we propose a novel \textit{weight-preserved loss} (WP Loss) for training the LAE to enhance the compression and reconstruction of large weights.~The WP Loss complements the original reconstruction loss while specifically prioritizing the preservation of large weights, and is mathematically expressed as:
\begin{equation}
    \begin{aligned}
        L_{\textit{WP}} = \frac{1}{n} \sum_{i=1}^{n} \left| x_i \right| \cdot \left| x_i - \hat{x}_i \right|,
    \end{aligned}
\end{equation}
where $n$ is the number of parameters, $x_i$ represents the $i$-th parameter, and $\hat{x}_i$ represents the $i$-th reconstructed parameter. The term $\left| x_i \right| \cdot \left| x_i - \hat{x}_i \right|$ emphasizes the reconstruction error for large weights. By integrating the WP Loss into the training process, our approach preserves more precise information about the identity in the large weights, thereby significantly improving the quality of generated images post-reconstruction. Moreover, our LAE achieves compression of the LoRA weights by nearly \textbf{300} times without compromising reconstruction performance.

\subsection{Diffusion Process}
\label{diffusion process} 

\hspace{1em} Transformers have demonstrated exceptional capabilities in handling long sequences in the language domain \cite{erkocc2023hyperdiffusion}, making them an ideal choice for modeling the LoRA latent representations. Based on the diffusion transformer (DiT) architecture, our diffusion model processes noisy LoRA latent representations and predicts their original forms.

During diffusion modeling, we optimize the parameters \(\theta\) of our model \( M_\theta \) to minimize the loss function:
\begin{equation}
    \begin{aligned}
        L(\theta) = \mathbb{E}_{t, x_0, c} \left[ w_t \left\| M_\theta (\alpha_t x_0 + \sigma_t \epsilon, t, c) - x_0 \right\|^2 \right],
    \end{aligned}
\end{equation}
here, \( M_\theta \) denotes the DiT model parameterized by \(\theta\), \( t \) is a time step uniformly sampled from the range \([0, T]\), and \( \epsilon \) is standard Gaussian noise. The parameters \( \alpha_t \), \( \sigma_t \), and \( w_t \) are diffusion noise parameters. The term \( x_0 \) represents the original latent representations encoded by LAE, while \( c \) represents the conditioning signal, obtained through MIF and integrated via Adaptive Layer Normalization (AdaLN) mechanisms to guide the diffusion process.

The training of the diffusion process and inference stage are illustrated in \cref{fig:train-inference}(b) and \cref{fig:train-inference}(d), respectively. During inference, DiffLoRA transforms random noise into LoRA latent representations through a guided denoising process, where the reference image provides conditioning information. We employ the DDIM \cite{ddim} strategy to sample new LoRA weights during this stage.

\subsection{Mixed Image Features (MIF)}
\label{mif}
\hspace{1em} The core concept of MIF is to leverage both facial details and general image information for enhanced identity feature extraction, thereby improving the performance of the denoising process. Drawing inspiration from Mixture-of-Experts (MoE) architecture \cite{shazeer2016outrageously}, MIF employs a gate network to dynamically combine face features and image features into the mixed image features. The overall workflow of MIF module is illustrated in \cref{fig:train-inference}(c).

Given a reference image \textit{I}, we extract two types of embeddings: facial details from the InsightFace encoder \(E_{\textit{face}}\) \cite{insightface} for identity recognition, and overall appearance features from the CLIP image encoder \(E_{\textit{img}}\) \cite{radford2021learning}. These embeddings are concatenated to form the mixed embedding:
\begin{equation}
    \begin{aligned}
        M = \textit{Concat}(E_{\textit{face}}(I), E_{\textit{img}}(I)),
    \end{aligned}
\end{equation}

By passing the mixed embedding \(M\) to the Gate Network \(G\), we compute importance scores \(S_{\textit{face}}\) and \(S_{\textit{img}}\), which represent the relative contributions of face features and image features respectively:

\begin{equation}
    \begin{aligned}
        S_{\textit{img}}, S_{\textit{face}} = G(M) = \textit{Softmax}(f(M)),
    \end{aligned}
\end{equation}
where \(f\) represents a linear transformation layer. 

The final mixed features \(F\) are computed as a weighted combination of the transformed embeddings:
\begin{equation}
        F = S_{\textit{img}} \odot MLP(E_{\textit{img}}(I)) + S_{\textit{face}} \odot MLP(E_{\textit{face}}(I)),
\end{equation}
where \(\odot\) denotes element-wise multiplication, and MLP represents the multi-layer perceptron that transforms each embedding into a compatible feature space.

\subsection{LoRA Weights Pipeline}
\label{pipeline}
\begin{figure}[htbp]
    \centering
    \includegraphics[width=\linewidth]{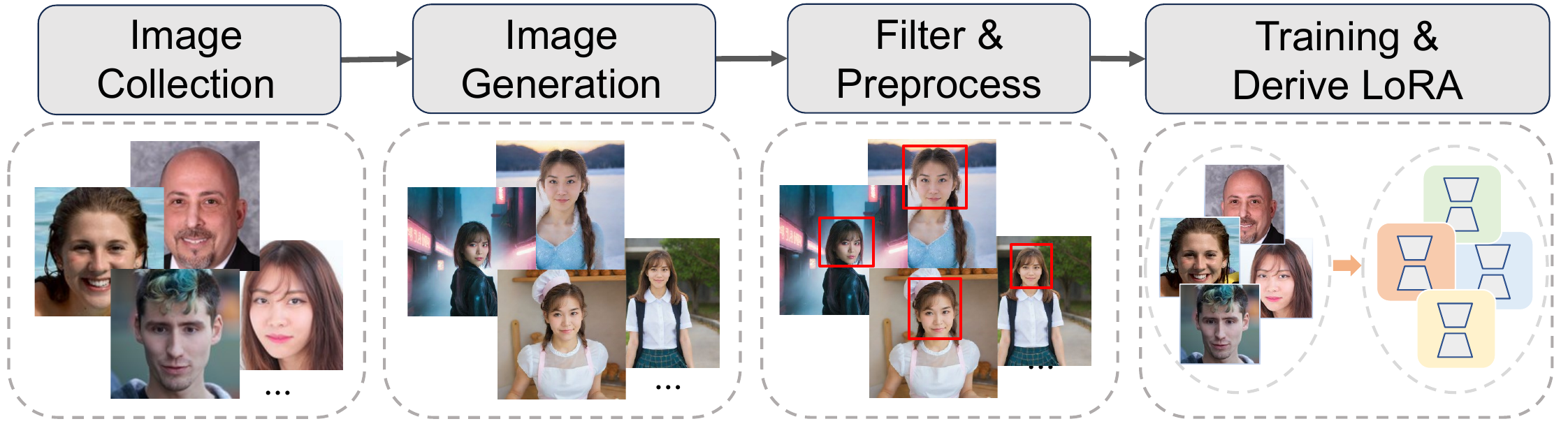}
    \caption{Pipeline for constructing LoRA weights dataset.
    }
\label{fig:dataset_pipeline}
\end{figure}

\hspace{1em} As illustrated in \cref{fig:dataset_pipeline}, we establish a comprehensive pipeline for generating high-quality LoRA weights datasets. This pipeline enables us to construct a diverse identity image dataset, where each identity encompasses 85 images, and then obtain LoRA weights for each individual by training with LoRA-DreamBooth \cite{von-platen-etal-2022-diffusers}, ultimately yielding 100k LoRA checkpoints.~The detailed implementation of our pipeline is presented in the Supplementary Materials.

\noindent\textbf{Image Collection.}~We begin by collecting high-quality facial images from the FFHQ \cite{FFHQ} and CelebA-HQ \cite{CelebA-HQ} datasets, ensuring diverse representation across different ages, ethnicities, and facial characteristics.

\label{experiments}
\begin{figure*}[htbp]
  \centering
  \includegraphics[width=\linewidth]{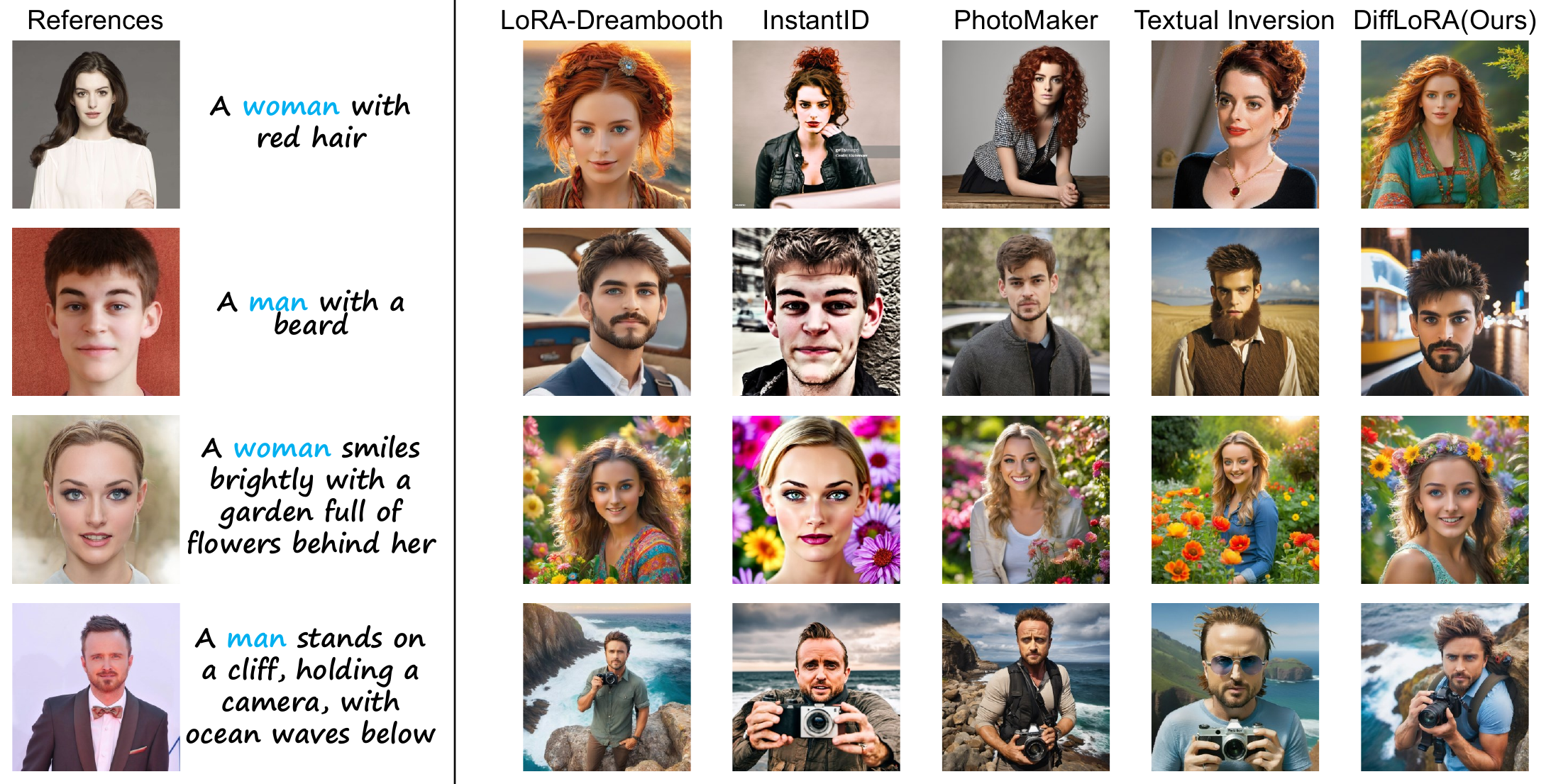}
  \caption{
    Qualitative comparison with several baseline methods.
  }
  \label{fig:experiment}
\end{figure*}
\begin{table*}[htbp]
  \center  
  \fontsize{10}{11.2}\selectfont
  \begin{tabular}{lccccccc}
    \toprule
    \textbf{Method} & \textbf{Image Reward$\uparrow$} & \textbf{DINO$\uparrow$} & \textbf{CLIP-I$\uparrow$} & \textbf{CLIP-T$\uparrow$} & \textbf{Face Sim$\uparrow$} & \textbf{FID$\downarrow$} & \textbf{Speed$\downarrow$(s)} \\
    \midrule
    LoRA-DreamBooth\cite{von-platen-etal-2022-diffusers} & 1.36 & 46.0 & 67.9 & 31.5 & 39.6 & 438.2 & 1052 \\
    Textual Inversion\cite{gal2022image} & 1.21 & 32.3 & 67.3 & 31.7 & 31.8 & 451.2 & 762 \\
    InstantID\cite{wang2024instantid} & 0.54 & 42.6 & 68.1 & 28.9 & 42.9 & 448.1 & 36 \\
    PhotoMaker\cite{li2024photomaker} & 1.32 & 36.7 & 66.9 & \textbf{32.3} & 30.9 & 447.8 & 26 \\
    DiffLoRA (Ours) & \textbf{1.44} & \textbf{47.6} & \textbf{68.5} & 31.2 & \textbf{43.3} & \textbf{437.4} & \textbf{20} \\    
    \bottomrule
\end{tabular}
  \caption{Comparison of DiffLoRA with several baselines across metrics: identity fidelity (DINO, CLIP-I, Face Sim), text-image consistency (CLIP-T, Image Reward), image quality (FID), and inference cost (personalization speed).}
  \vspace{-1em}
  \label{tab:experiment}
\end{table*}

\begin{table}[htbp]
    \centering
    \setlength{\tabcolsep}{3pt}
    \fontsize{8}{11.2}\selectfont
    \begin{tabular}{lcccc}
      \toprule
      \textbf{Method} & \textbf{Fidelity$\uparrow$} & \textbf{Plausibility$\uparrow$} & \textbf{Alignment$\uparrow$} & \textbf{Quality$\uparrow$} \\
      \midrule
      Textual Inversion\cite{gal2022image} & 2.24 & 3.92 & 3.88 & 3.64 \\
      InstantID\cite{wang2024instantid} & \textbf{4.15} & 2.01 & 2.32 & 3.61 \\
      PhotoMaker\cite{li2024photomaker} & 3.79 & 3.86 & 3.81 & 4.12 \\
      Ours & 3.94 & \textbf{4.21} & \textbf{4.30} & \textbf{4.22} \\
      \bottomrule
    \end{tabular}
    \caption{User study results across the perspectives of Fidelity, Plausibility, Alignment, and Quality.}
    \vspace{-1em}
    \label{tab:userstudy}
  \end{table}
  
\noindent\textbf{Image Generation.} Utilizing PhotoMaker \cite{li2024photomaker}, we first generate 100 images for each individual with 100 specifically crafted prompts. These prompts are gender-specific and designed to capture diverse perspectives, expressions, attributes, and scenes, ensuring comprehensive coverage of each identity's characteristics.

\noindent\textbf{Image Filtering and Preprocessing.} We employ InsightFace \cite{insightface} to compute facial similarity between the generated images and the original image. The top 85 images with the highest similarity scores are selected and subsequently augmented through data preprocessing operations including cropping and flipping to prevent overfitting.

\noindent\textbf{Personalized LoRA Training.}~The 85 preprocessed images are utilized to fine-tune SDXL through LoRA-DreamBooth \cite{von-platen-etal-2022-diffusers}, deriving specialized LoRA weights that enable high-fidelity personalized image generation while following the optimization strategies described in \cite{yeh2023navigating}.

\section{Experiments}  

  \subsection{Implementation Details} 

  \hspace{1em} Based on the identity images and high-quality LoRA weights obtained from \cref{pipeline}, we implement DiffLoRA as follows. We employ a 5-layer 1D convolutional encoder in LAE to compress the flattened LoRA weights vectors into latent representations.~Each layer contains two residual blocks, with an attention layer in the middle block. For the decoder, we utilize attention layers combined with an MLP to reconstruct the LoRA latent representations. We optimize LAE with AdamW \cite{kingma2014adam} on 4 NVIDIA A6000 GPUs for 600 GPU hours with a batch size of 4 and a learning rate of 2e-5. For the diffusion process, we train a 16-layer DiT model with 1454 dimensions. During training, we randomly sample images of the same identity as reference images to create mixed image features. We optimize the DiT model using AdamW with a batch size of 32 and a learning rate of 1e-4. The diffusion process implements 1000 timesteps with a linear noise scheduler ranging from 0.0001 to 0.012. We train the model for 1400 epochs, requiring about 300 GPU hours on 4 NVIDIA A6000 GPUs. For inference, we use a single reference image and generate personalized LoRA weights using 100 steps of the DDIM sampler.

\subsection{Evaluation Metrics} 
\hspace{1em} Following DreamBooth \cite{ruiz2023dreambooth}, we use  DINO \cite{caron2021emerging} and CLIP-I \cite{gal2022image} metrics to measure the \textit{identity fidelity}, and CLIP-T metric to assess \textit{text-image consistency}. However, the CLIP-T metric may not fully capture the nuances of text-image alignment. Therefore, we also compute the Image Reward score \cite{xu2024imagereward} to provide a more comprehensive evaluation of the generated images relative to the text prompts. Additionally, to further evaluate identity fidelity, we assess \textit{face similarity}, termed Face Sim, by calculating the embedding similarity between the generated and reference images using InsightFace. The \textit{image quality} is assessed using the FID metric \cite{heusel2017gans}, which compares the distribution of generated images with those from MS-COCO \cite{lin2014microsoft}. Lower FID values indicate better image quality. Furthermore, We assess the inference cost by evaluating personalization speed based on whether the method involves tuning or is tuning-free, providing a comprehensive evaluation of the end-to-end personalization. For tuning-free methods, only inference time is measured, whereas for methods requiring tuning, both training and inference times are included.

\subsection{Evaluation Dataset} 
\hspace{1em} Our evaluation dataset consists of 50 identities, evenly split between the SFHQ dataset \cite{sfhq} and our own collection. All reference images are disjoint from the training set to ensure an unbiased evaluation of the model's generalization ability. Additionally, we prepare a total of 30 prompts, encompassing simple, complex, and multi-angle prompts to ensure a comprehensive evaluation. For quantitative evaluation, we synthesized a set of 4 images for each prompt corresponding to each subject identity.

\subsection{Comparisons}
\hspace{1em} To evaluate the performance of DiffLoRA, we conducted a comparative analysis with several leading open-source techniques, including LoRA-DreamBooth \cite{von-platen-etal-2022-diffusers}, Textual Inversion \cite{gal2022image}, InstantID \cite{wang2024instantid}, and PhotoMaker \cite{li2024photomaker}. We employ the default settings as proposed in their respective works. For a fair comparison, we use SDXL versions for LoRA-DreamBooth and Textual Inversion, with LoRA-DreamBooth configured according to \cref{pipeline}. More results are listed in the Supplementary Materials.

\noindent\textbf{Qualitative Comparisons.} As illustrated in \cref{fig:experiment}, DiffLoRA consistently outperforms these methods by generating images with greater diversity and improved aesthetic quality. This is especially apparent in scenarios where InstantID demonstrates overfitting to the reference image, and PhotoMaker slightly compromises identity fidelity.

\begin{figure}[t]
  \centering
  \includegraphics[width=\linewidth]{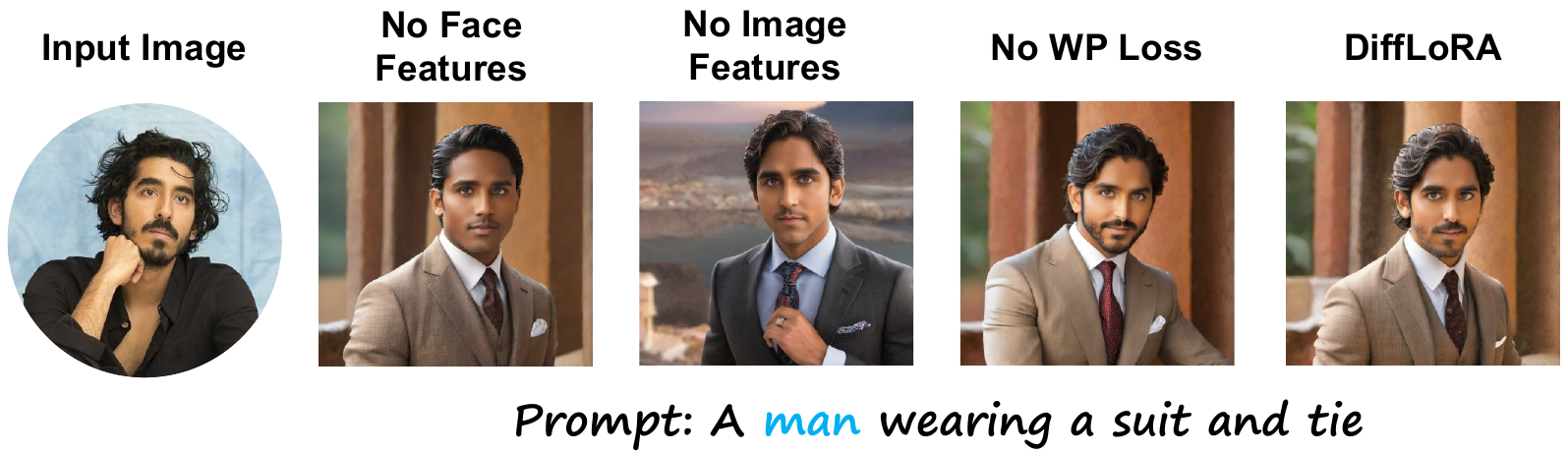}
  \captionsetup{width=1.02\linewidth}
  \caption{Visualization of generated results under different settings.}
  \label{fig:ablation}
\end{figure}

\begin{table}[t]
  \center 
  \setlength{\tabcolsep}{5pt}
  \fontsize{8.2}{11}\selectfont
  \begin{tabular}{lccc}
    \toprule
    \textbf{Method} & \textbf{Face Sim$\uparrow$} & \textbf{DINO$\uparrow$} & \textbf{CLIP-I$\uparrow$} \\
    \midrule
    No Face Feature & 20.42 & 42.13 & 63.10 \\
    No Image Feature & 39.25 & 45.69 & 64.29 \\
    No WP Loss & 36.38 & 47.26 & 60.18 \\
    Ours & \textbf{43.37} & \textbf{47.63} & \textbf{68.54} \\
    \bottomrule
  \end{tabular}
  \caption{Ablation studies on identity fidelity metrics (Face Sim, DINO, CLIP-I), Our full method performs best across all metrics.}
  \vspace{-1em}
  \label{tab:ablation}
\end{table}

\noindent\textbf{Quantitative Comparisons.} In our quantitative experiments, the effectiveness and efficiency of DiffLoRA were evaluated using multiple metrics covering various aspects: text-image consistency, identity fidelity, generation quality, and personalization speed. As depicted in \cref{tab:experiment}, our method achieved the highest scores in CLIP-I, DINO, and Face Sim metrics, demonstrating its capability to generate high-quality images with superior identity fidelity. Additionally, DiffLoRA exhibited the lowest inference cost compared to other methods. This indicates that DiffLoRA successfully balances high-quality image generation with computational efficiency.

\begin{figure*}[htbp]
  \vspace{-1em}
  \centering
  \begin{subfigure}{0.69\textwidth}
    \vspace{-1em}
    \centering
    \includegraphics[width=\linewidth]{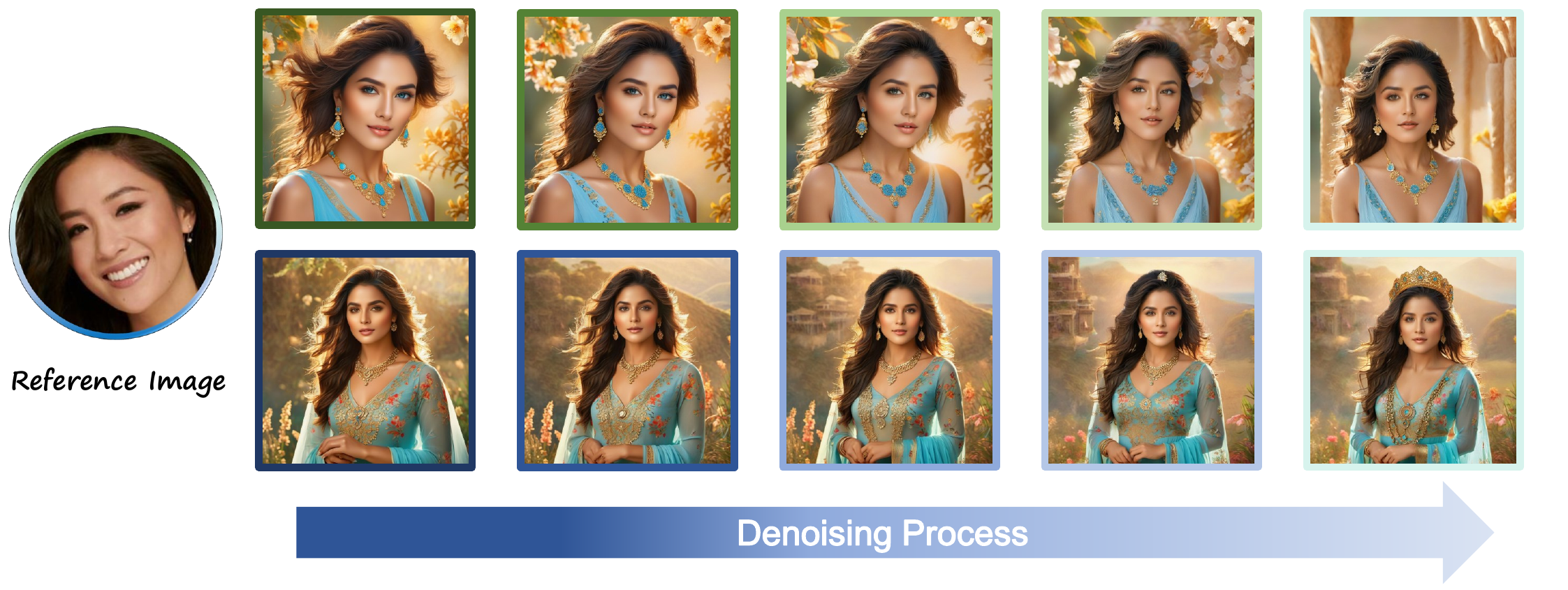}
    \caption{Visualization of the Denoising Process}
    \label{fig:denoising}
  \end{subfigure}
  \hfill
  \begin{subfigure}{0.3\textwidth}
    \vspace{-1em}
    \centering
    \includegraphics[width=\linewidth]{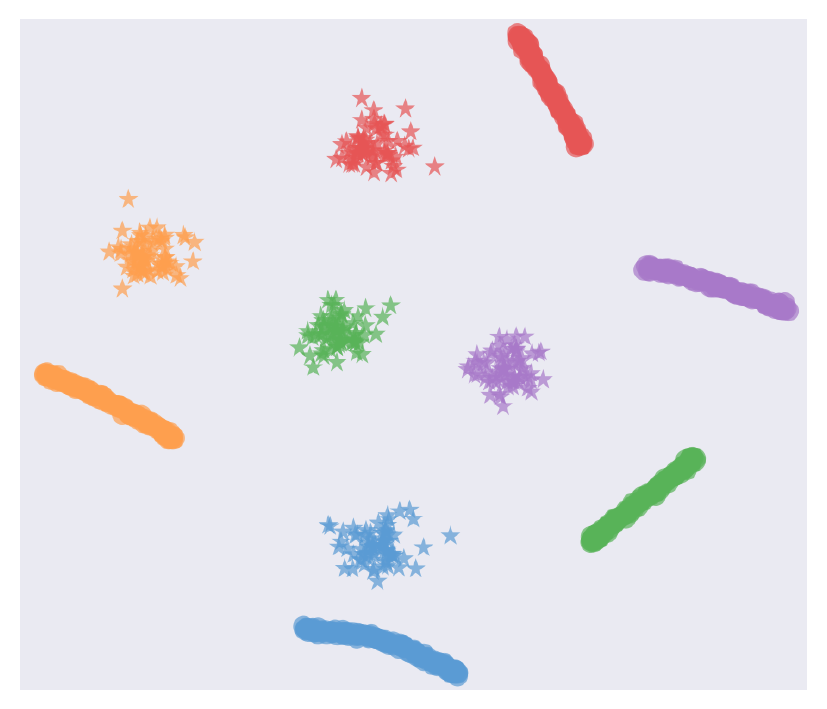}
    \caption{t-SNE Visualization}
    \label{fig:t-sne}
  \end{subfigure}
  \vspace{-0.75em}
  \caption{Visualization of the denoising process through progressive image generation at intermediate stages using personalized LoRA weights, and t-SNE comparison where \(\bigstar\) represents generated weights and \(\bullet\) represents training weights across five identities.}
  \vspace{-1.2em} 
\end{figure*}

\noindent\textbf{User Study.}~We conduct a user study to compare DiffLoRA with Textual Inversion, InstantID, and PhotoMaker. The study involved 20 participants evaluating 50 text-image pairs, where these pairs were distributed across multiple reference identities. Each pair was evaluated for all four methods, resulting in a total of 4,000 valid votes. We prepare detailed regulations and rating criteria. Each image is rated for scores of 1 to 5 from four perspectives: \textit{Fidelity} measures the ability of preserving identity features, \textit{Plausibility} evaluates the anatomical plausibility of the generated poses, \textit{Alignment} assesses the alignment with given text descriptions, and \textit{Quality} evaluates the overall visual fidelity and aesthetic coherence. As shown in \cref{tab:userstudy}, DiffLoRA outperforms other methods in Plausibility, Alignment, and Quality. InstantID achieves marginally higher Fidelity due to its tendency to overfit to reference images, yet it demonstrates substantially lower performance across other evaluation criteria. Our method maintains competitive Fidelity while excelling in other aspects, demonstrating a better balance between identity preservation and generation diversity.

\subsection{Ablation Studies}
\hspace{1em} As illustrated in \cref{tab:ablation} and \cref{fig:ablation}, our analysis reveals the impact of different components of the DiffLoRA method on the quality of generated images. 

\noindent\textbf{Impact of the Mixed Image Features (MIF).} When the gate network in MIF is removed, using only face features (No Image Feature) or only image features (No Face Feature) for inference, there is a significant decrease in the identity preservation of the generated images. This drastic drop suggests that MIF plays a crucial role in guiding DiffLoRA to generate accurate LoRA weights. Especially, when relying solely on image features, the identity features of personalized generation are nearly lost.

\noindent\textbf{Impact of the Weight-Preserved Loss (WP Loss).} We train the LAE without the WP Loss (No WP Loss), which results in a noticeable decline in image quality. The generated identity is less similar to the input identity without WP Loss. Incorporating the WP Loss significantly improves the LAE's capacity to retain detailed identity information within the weights. Consequently, the Face Sim score increases from 36.38 to 43.37.

\subsection{Analysis}
\textbf{Superior Weights Generation via Diffusion Models.} As demonstrated in \cref{fig:experiment} and \cref{tab:experiment}, DiffLoRA consistently outperforms LoRA-DreamBooth, despite the latter's alignment with the pipeline in \cref{pipeline}. We attribute this phenomenon to the diffusion model's unique probabilistic modeling and its comprehensive exploration of the parameter space. By representing data distributions in a high-dimensional latent space, the diffusion model transcends the limitations of original weights, identifying optimal solutions beyond the training data and avoiding local minima \cite{sohl2015deep, ddpm}, thereby capturing deeper structures and patterns.

To further substantiate our claims,~we employ t-SNE \cite{van2008visualizing} to visualize the distributions of weights obtained from consecutive steps in \cref{pipeline} and those generated by DiffLoRA across multiple identities, as shown in \cref{fig:t-sne}. The distinct distributions indicate broader weight space exploration and enhanced generalization.s~Additionally, \cref{fig:denoising} illustrates the denoising process of personalized LoRA weights, showing that the diffusion model initiates from stochastic initial conditions, progressively refines the weights, and incrementally converges towards the target distribution.~This process aligns with the information bottleneck theory \cite{tishby2000information}, which states that by retaining only the most relevant features and eliminating redundant information, the model improves generalization and mitigates overfitting. Furthermore, this process facilitates optimization by smoothing the loss landscape, which allows the model to discover more favorable local minima or global optima \cite{neyshabur2018towards},~ultimately leading to more expressive weight configurations.

\noindent\textbf{Generalization Capability.} Since LoRA is a general fine-tuning method, DiffLoRA naturally inherits its cross-model and cross-task adaptability. To validate the versatility of our method, we implemented DiffLoRA on the SD 1.5 \cite{rombach2022high} for style transfer tasks. Specifically, we successfully generate different stylistic LoRA weights (e.g., photorealistic, anime, and sketches) based on the reference images. More examples are shown in the Supplementary Materials.

\section{Conclusion}

\hspace{1em} In this work, we present DiffLoRA, a novel method for human-centric personalized image generation.~Our~approach~employs~a latent diffusion-based hypernetwork framework to predict LoRA weights for specific identity adaptation in the SDXL,~enabling tuning-free generation of high-fidelity portraits without extra inference cost.~Experimental~results~show~that~DiffLoRA~outperforms~existing methods~in~text-image~consistency,~identity fidelity,~generation quality,~and inference efficiency.~To~the~best~of~our~knowledge,~we~make~the~first~attempt~to~leverage~diffusion~models~for~personalized~LoRA~weights~generation,~paving~the~way~for~more adaptive frameworks~in~diverse~architectures.

{
    \small
    \bibliographystyle{ieeenat_fullname}
    \bibliography{main}
}

\end{document}